\documentclass[11pt]{article}
\pdfoutput=1

\usepackage[margin=1in]{geometry}
\usepackage{amsmath,amssymb}
\usepackage{multirow}
\usepackage{float}
\usepackage{graphicx}
\usepackage{authblk}
\usepackage[numbers]{natbib}
\usepackage{url}
\usepackage{microtype}

\AtBeginDocument{%
  
}

\title{The Impact of Data Characteristics on \\GNN Evaluation for Detecting Fake News}

\author[1]{Isha Karn}
\author[1]{David Jensen}
\affil[1]{University of Massachusetts Amherst, Amherst, MA, USA\\
\texttt{\{ikarn,jensen\}@umass.edu}}

\date{} % remove date; or set a specific date if you like

\begin{document}

\maketitle

\begin{abstract}
Graph neural networks (GNNs) are widely used for the detection of fake news by modeling the content and propagation structure of news articles on social media. We show that two of the most commonly used benchmark data sets --- GossipCop and PolitiFact --- are poorly suited to evaluating the utility of models that use propagation structure. Specifically, these data sets exhibit shallow, ego-like graph topologies that provide little or no ability to differentiate among modeling methods. We systematically benchmark five GNN architectures against a structure-agnostic multilayer perceptron (MLP) that uses the same node features. We show that MLPs match or closely trail the performance of GNNs, with performance gaps often within 1-2\% and overlapping confidence intervals. To isolate the contribution of structure in these datasets, we conduct controlled experiments where node features are shuffled or edge structures randomized. We find that performance collapses under feature shuffling but remain stable under edge randomization. This suggests that structure plays a negligible role in these benchmarks. Structural analysis further reveals that over 75\% of nodes are only one hop from the root, exhibiting minimal structural diversity. In contrast, on synthetic datasets where node features are noisy and structure is informative, GNNs significantly outperform MLPs. These findings provide strong evidence that widely used benchmarks do not meaningfully test the utility of modeling structural features, and they motivate the development of datasets with richer, more diverse graph topologies.
\end{abstract}


\section{Introduction}
Graph Neural Networks (GNNs) have become a dominant modeling paradigm in machine learning, especially for tasks involving relational data such as social networks, citation graphs, and protein structures. Among these, misinformation detection on social media has emerged as a prominent application, where GNNs are used to integrate textual signals with the structure of information propagation. However, as GNNs grow in popularity, it becomes increasingly important to critically assess whether their performance gains genuinely stem from leveraging graph structure or whether they are primarily driven by highly informative node features. In this paper, we visit this question through the lens of detecting fake news on two widely used benchmark datasets: PolitiFact and GossipCop.

These two datasets, first introduced in the FakeNewsNet repository \cite{shu2020fakenewsnet}, have become foundational benchmarks in fake news detection, collectively cited in more than 1,500 studies across modalities such as graph neural networks, transformers, and multi-modal learning. Their popularity comes from the combination of article content and associated social graphs, making them a natural fit for models that integrate structure and features. However, this same combination can mask whether structure actually contributes to performance.

Despite their wide use, there has been little scrutiny of the assumptions these datasets encode. We find that both consist largely of shallow, ego-like graphs centered on a single news article, with minimal structural complexity. At the same time, their node features, which are derived from BERT and spaCy, are often linearly separable by class. This leads us to ask a critical question:

\begin{quote}
\textit{To what extent are GNNs learning from graph structure versus relying only on the information implicit in node features?}
\end{quote}

This question addresses a key issue for the evaluation of GNNs. If strong performance can be achieved without meaningful structure, then benchmark results may conflate representational power with architectural necessity. While prior work uses PolitiFact and GossipCop to evaluate GNN-based models, we argue that these datasets are inadequate for evaluating models that are expected to exploit both structural and feature-based signals. Their highly ego-like topology and strongly separable node embeddings limit the ability to assess whether models genuinely benefit from relational reasoning.

We approach this problem through a comprehensive evaluation of both real-world and synthetic graph datasets. On the real datasets, we evaluate five GNN architectures alongside a multilayer perceptron (MLP) baseline that uses the same node features but no graph structure. On the synthetic side, we design datasets that allow us to control the contribution of node features and structural patterns. To further isolate the role of structure, we conduct controlled perturbation experiments by randomly shuffling node features and rewiring edges. These manipulations reveal that model performance collapses for the most part under feature shuffling but remains stable under edge shuffling: indicating that structure plays a negligible role in classification on these benchmarks.

These experiments reveal that GNNs only provide consistent advantages when node features are weak and structural signals are non trivial. In contrast, when node embeddings are already linearly separable --- as is in the case of GossipCop and PolitiFact --- GNNs perform only marginally better than MLPs, and sometimes not at all. Together these results suggest that high benchmark performance does not reflect structural reasoning, but instead hinges on informativeness of pretrained node features. If node features are already informative, GNNs may act primarily as amplifiers, leading us to incorrectly attribute performance. For instance, on GossipCop, an MLP trained on node features alone achieves accuracy on par with GCN-based models. Only when feature-label alignment is deliberately weakened through shuffling or noise, do GNNs show a meaningful advantage. This has implications for how future datasets are designed and how GNN-based systems are evaluated.

\section{Background}
We begin by reviewing key components of the task: fake news detection, the role of GNNs in solving this task, and the datasets and practices used to evaluate GNN-based systems.

\subsection{Fake News Detection}
Fake news detection has been approached through various classes of methods, including content-based, social context-based, and multimodal methods. Early work relied heavily on textual features such as sentiment, syntax, and source metadata \cite{figueira2017current, katarya2020recognizing, khan2022detecting, raza2022fake}. These approaches were often paired with traditional classifiers such as logistic regression or support vector machines, \cite{zhou2020survey, shu2017fake} and limited by their dependence on surface-level features and labeled data \cite{orabi2020detection}. More recent approaches incorporate user metadata, temporal activity, or external knowledge sources such as knowledge graphs \cite{dun2021kan}.
Many methods attempt to enhance detection performance through multimodal cues, including visual features \cite{zhou2020safe} or explanatory signals \cite{shu2019defend}. These models show strong performance on benchmark datasets such as PolitiFact and Gossipcop, achieving accuracies exceeding 85\% on PolitiFact and over 90\% on GossipCop in some cases. However, these results often come from pipelines that tightly couple textual signals with high-capacity encoders such as BERT, leaving the role of structural information underexplored. 

\subsection{Graph Neural Networks for Fake News Detection}

To capture social propagation dynamics, many recent studies have shifted toward graph-based models. These models represent tweets, users, or posts as nodes and use retweet, reply, or follow relationships as edges, forming a social context graph. GNNs such as GCN \cite{kipf2016semi}, GAT \cite{velickovic2017graph}, and GraphSAGE \cite{hamilton2017inductive} have been adopted to encode these networks. 

A range of specialized GNN-based architectures have since been proposed, including UPFD \cite{dou2021user}, which introduces attention-based user preference modeling; GAMC \cite{yin2024gamc}, which incorporates masking and contrastive learning to improve robustness; PSGT \cite{zhu2024propagation}, which applies graph transformers for interpretable propagation modeling; and GAT/GCN/SAGE-GANM \cite{chang2024graph}, which leverages global attention and memory for enhanced structural reasoning. \cite{jung2023topological} introduced a two-state GNN with temporal encoding for modeling sequential post activity; \cite{saikia2022modelling} and \cite{wu2023decor} explored hybrid GNN-transformer models that integrate both textual and structural features; and \cite{davoudi2022dss} proposed propagation-based models for early-stage detection. These systems consistently report strong performance, often surpassing LSTM or CNN-based baselines, with GAT-based models achieving over 97\% accuracy on GossipCop. 

\subsection{Benchmark Datasets and the Role of Structure}

The \textit{PolitiFact} and \textit{GossipCop} datasets, released as part of the FakeNewsNet repository \cite{shu2020fakenewsnet}, have become two of the most widely used benchmarks for fake news detection. Their popularity stems from their hybrid design: each instance includes both textual content (e.g., claims and tweets) and social graph capturing propagation dynamics on twitter. This makes them appealing for models that combine content-based features with relational signals, especially GNNs.

As observed above, GNN-based models report strong performance on these datasets and often attribute their success to the incorporation of social context. However, these evaluations are typically end-to-end and relatively few studies attempt to disentangle the contribution of structure from that of node features. Some prior work\cite{nguyen2020fang, jeong2022nothing, barnabo2022fbmultilingmisinfo} highlights issues such as missing metadata, dataset obsolescence, or noisy nodes, but none systematically examine whether the structure encoded in these datasets meaningfully contributes to GNN performance.

A closer inspection reveals that these benchmarks consist largely of shallow, ego-like graphs centered on a single news post, with limited relational diversity. Node features, often derived from pretrained language models like BERT tend to be highly informative and linearly separable. So, even simple models that ignore structure (e.g., MLPs) may match or approach the performance of GNNs. These conditions risk inflating the perceived effectiveness of structural reasoning in GNNs for the detection of fake news, even when little actual structural signal is being exploited.

Surprisingly, we find little evidence in literature that directly compares GNN-based architectures to structure-agnostic baselines trained on the same features. Nor are there many ablations studies that isolate the effect of edges or evaluate performance under controlled synthetic conditions. As a result, the validity of GossipCop and PolitiFact as benchmarks for structural reasoning remains under-examined.

In this work, we revisit these assumptions through a broader lens, evaluating multiple GNN-based architectures and compare them against MLP baselines trained on identical node features. To precisely isolate the contribution of structure, we conduct controlled perturbation experiments in which we shuffle node features or randomly rewire edges. We also conduct synthetic graph datasets where feature and structure signals can be independently varied. This dual approach allows us to identify not just whether GNNs outperform simpler models, but under what conditions structural reasoning meaningfully contributes to performance.

\section{Method}

\subsection{Data}
\subsubsection{Real-World Datasets}

We use the PolitiFact and GossipCop datasets from the FakeNewsNet repository \cite{shu2020fakenewsnet, dou2021user}. Each instance in these datasets represents the propagation of a news article on X (formerly Twitter) as a graph. The root node denotes the original news article, while the leaf nodes correspond to who retweeted or replied to it. Edges capture these retweet and reply relationships, resulting in shallow, star-shaped ego-graphs centered on a single post.
Each node is associated with several feature sets: (1) 10-dimensional user profile metadata; (2) 300-dimensional spaCy embeddings of post content; (3) 768-dimensional BERT embeddings; and (4) 310-dimensional manually extracted content features. We clarified that the BERT and spaCy embeddings are computed independently from graph structure and that no propagation or interaction signal is embedded in the features themselves. These features are fixed across all experiments to isolate the effect of graph structure. Despite differences in size and topology, both datasets are approximately balanced between fake and real news classes. Summary statistics are provided in Table \ref{tab:graph_stats}.

To test model sensitivity to input quality and structural signal, we apply two forms of perturbation across both real-world and synthetic settings. First, in the feature shuffling setting, node features are randomly permuted across the entire dataset, reassigning them arbitrarily to different graphs. This process destroys the marginal distribution of node features within each graph, while approximately preserving the joint distribution of feature vectors across the dataset. This allows us to test whether models rely on intra-graph feature coherence or can still succeed when structural alignment with features is broken. Second, in the shuffled edges setting, graph edges are rewired at random while node features and labels are kept intact. These perturbations allow us to isolate whether models are leveraging structure, features, or both. 

\begin{table}[t]
\caption{Dataset and graph statistics for GossipCop and PolitiFact. The number of fake vs. real graphs is balanced in both datasets.}
\label{tab:graph_stats}
\centering
\begin{tabular}{|l|l|l|l|l|}
\hline
Dataset &
  \#Graphs &
  \begin{tabular}[c]{@{}l@{}} \#Nodes\end{tabular} &
  \begin{tabular}[c]{@{}l@{}} \#Edges\end{tabular} &
  \begin{tabular}[c]{@{}l@{}}Avg. Nodes \\ per Graph\end{tabular} \\ \hline
GossipCop &
  5,464 &
  314,262 &
  308,798 &
  58 \\ \hline
PolitiFact &
  314 &
  41,740 &
  40,740 &
  131 \\ \hline
\end{tabular}
\end{table}

\subsubsection{Synthetic Datasets}

To isolate the contributions of graph structure and node features, we construct synthetic ego-graphs under controlled conditions. Each graph belongs to one of two classes defined by their structural topology: Class 0 graphs follow a densely connected clique pattern, while Class 1 graphs adopt a star-like hub-and-spoke configuration. 

We generate graphs under three different feature regimes:
\begin{enumerate}
    \item \textbf{Clean Features:} Node features are drawn from class-specific multivariate Gaussian. The structure still retains predictive signal.
    \item \textbf{Structure Only}: Node features are i.i.d. random noise, leaving structure as the only source of predictive signal.
    \item \textbf{Noisy Features:} Node features are drawn from their correct class distribution with 80\% probability and corrupted with 20\% noise to simulate partial feature reliability.
\end{enumerate}

To further test whether models rely on structural reasoning or exploit direct correlations between features and labels, we introduce a feature randomization variant applied only to the clean features setting. In this variant, node features are randomly permuted across the dataset, breaking the alignment between features and structure while preserving the original graph topology. 
All synthetic datasets are class-balanced and use fixed-dimensional node features. Minor variation in graph size and connectivity is introduced to simulate natural heterogeneity while maintaining control over signal sources. 

\subsection{System}
\subsubsection{Models}

We evaluate a diverse set of GNN architectures and baselines to assess the contribution of structure for the task of detecting fake news. Our GNN models include:
\begin{enumerate}
    \item \textbf{GCN:} A Graph Convolutional Network that aggregates neighborhood features using symmetric normalization of the adjacency matrix. 
    \item \textbf{GAT:} A Graph Attention Network that learns attention weights over neighbors, enabling more expressive message passing by focusing on relevant nodes. 
    \item \textbf{GraphSAGE:} An inductive GNN that learns to aggregate sampled neighborhood information using mean pooling, making it scalable to unseen graphs. 
    \item \textbf{GCNFN:} A GAT-based model by \cite{monti2019fake} that encodes propagation structure via two attention layers and concatenates the resulting node embedding with precomputed textual features (e.g., BERT). The fused representation is passed through two dense layers for classification. This architecture is designed to combine content and structure in a single forward pass. 
    \item \textbf{GNNCL:} A propagation-based GNN model by \cite{han2020graph}, augmented with continual learning techniques. It uses Gradient Episodic Memory (GEM) and Elastic Weight Consolidation (EWC) to prevent catastrophic forgetting when adapting to new datasets. The model uses a DiffPool-based encoder for graph-level classification and is designed to retain performance across dataset shifts without retraining from scratch. 
\end{enumerate}

As a baseline, we include a multilayer perceptron (MLP) that uses all the node features for each instance but ignores graph structure entirely. For each graph, the MLP averages the node features and applies two fully connected layers to predict the label. 

\subsubsection{Synthetic Dataset Models}

For synthetic datasets, we evaluate four models: GCN, GAT, GraphSAGE, and MLP. Each GNN uses two message-passing layers followed by global mean pooling. The MLP aggregates node features by averaging and applies two dense layers. This setup enables direct comparison between structure-aware and structure-agnostic approaches under varying signal conditions.

\subsubsection{Training and Evaluation}

All models are implemented in PyTorch and PyTorch Geometric. For real-world datasets, we use the full 768-dimensional node features, a batch size of 128, and an Adam optimizer with a weight decay of 0.001. For synthetic datasets, node features are 32-dimensional and class-balanced by construction. All experiments use stratified 10-fold cross-validation, and we report mean validation accuracy with 95\% confidence intervals. We added that the same features are used across models and that we use stratified folds. All experiments were run on an NVIDIA RTX 4070 Mobile GPU.

\section{Results}

We evaluate the performance of several GNN architectures and a structure-agnostic MLP baseline on two real-world datasets: GossipCop and PolitiFact, as well as a suite of synthetic datasets. The goal is to isolate and quantify the relative contributions of node features and graph structure to the classification performance.

\subsection{Real-World Benchmarks: Do GNNs add value?}
Our experiments on GossipCop and PolitiFact show that GNNs are not always necessary to achieve high performance, as seen in Table \ref{tab:base_gos-pol}. On GossipCop, GCNFN and GraphSAGE achieve the highest accuracies (96.06\% and 95.54\%, respectively), but surprisingly, a simple MLP that ignores graph structure performs nearly as well, with 93.87\% accuracy. This suggest that node features alone are highly informative on the dataset, a trend clearly visible in the accuracy curves in Figure \ref{fig:gos_pol_shuffled}.

\begin{figure}[t]
    \centering
    \includegraphics[width=1\linewidth]{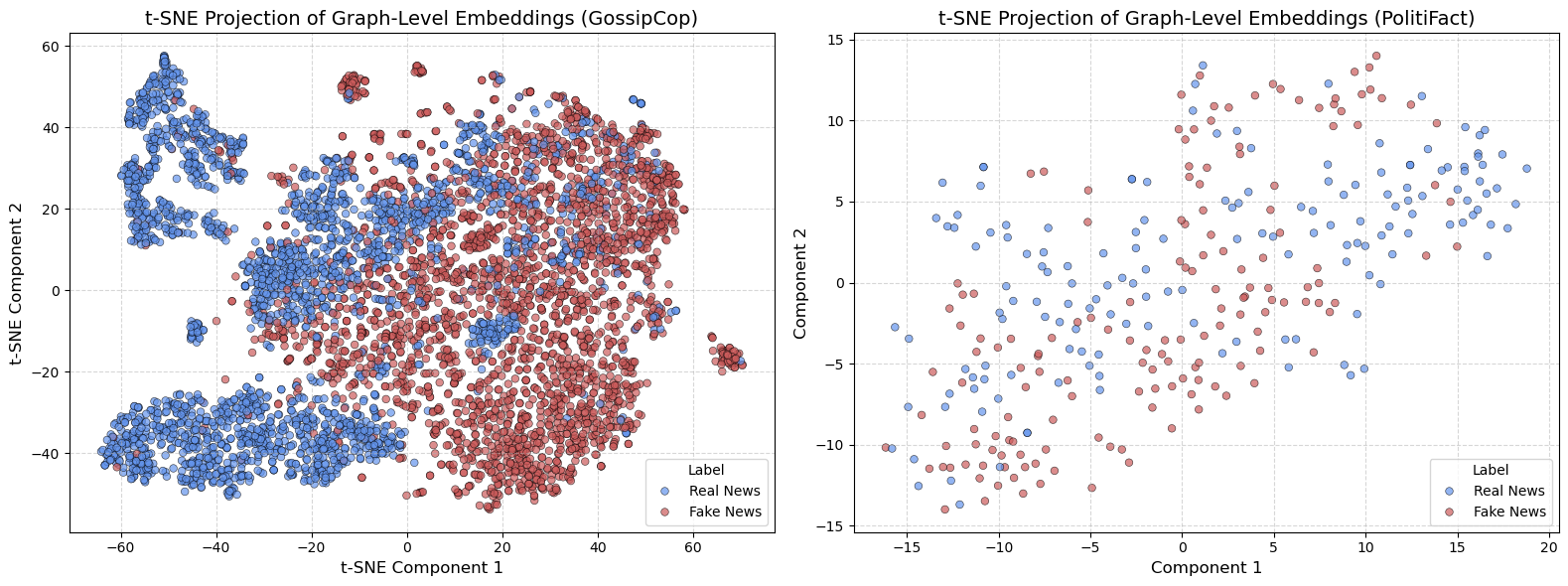}
    \caption{t-SNE Projection of node features for GossipCop (left) and PolitiFact (Right)}
    \label{fig:tsne}
\end{figure}

This trend aligns with our t-SNE visualizations of node embeddings, which show that real and fake news instances in GossipCop are already well-separated in feature space. As shown in Figure \ref{fig:tsne}, real and fake instances are linearly separable in feature space even before any structural modeling. GossipCop focuses on celebrity and entertainment content, which tends to include emotionally charged or stylized language. These cues are likely captured well by pretrained language models like BERT, reducing the need for structural modeling.

PolitiFact is a smaller dataset that focuses on political claims and fact-checks. Its nodes are also separable but due to its smaller size, there is more variance in all models. GCNFN reaches 82.71\% accuracy, while the MLP reaches 80.93\% accuracy. The performance differences fall within overlapping confidence intervals, suggesting again that structure contributes little relative to strong feature embeddings.

To test whether GNNs can use structural signals alone to correctly classify data instances, we conduct feature shuffling, where node features are randomly permuted across the dataset. This destroys the marginal distribution of features within graphs while preserving the overall joint distribution of node features across the dataset. Under this setting for GossipCop, we observe that GCN and GNNCL maintain performance at around 80\%, while GAT and GraphSAGE degrade to mere chance. This suggests that GCN and GNNCL may be leveraging dataset-levelo structural biases such as graph size or density, which are found to correlate with labels. GCN's global aggregation and GNNCL's hierarchical pooling appear better suited to capture these coarse structural biases. In contrast, attention-based or sampling-based methods such as GAT and GraphSAGE appear unable to exploit the characteristics of sparse, ego-like graphs where local neighborhoods provide limited information, especially when node features are no longer reliable.

Next, we apply structure randomization, where graph edges are randomly rewired while node features remain fixed. Unsurprisingly, performance remained strong across all models, including GCNFN, GAT, GraphSAGE, the structure-agnostic MLP, with accuracy exceeding 90\%. This indicates that structure is not essential to the decision process, and that node features dominate classification.

We observe a similar trend on PolitiFact under randomized edge structure. All models continue to improve steadily across epochs, with validation accuracies rising well above 70\% for most architectures. This reinforces the conclusion that structural information is not essential for high classification performance. While GAT shows the highest final accuracy in this setting, the difference from the MLP is minimal, suggesting that node features again dominate predictive power, even when the original graph connectivity is entirely disrupted. If anything, it further highlights the dominance of node features in driving classification performance on GossipCop and PolitiFact. 

\begin{figure}[t]
    \centering
    \includegraphics[width=1\linewidth]{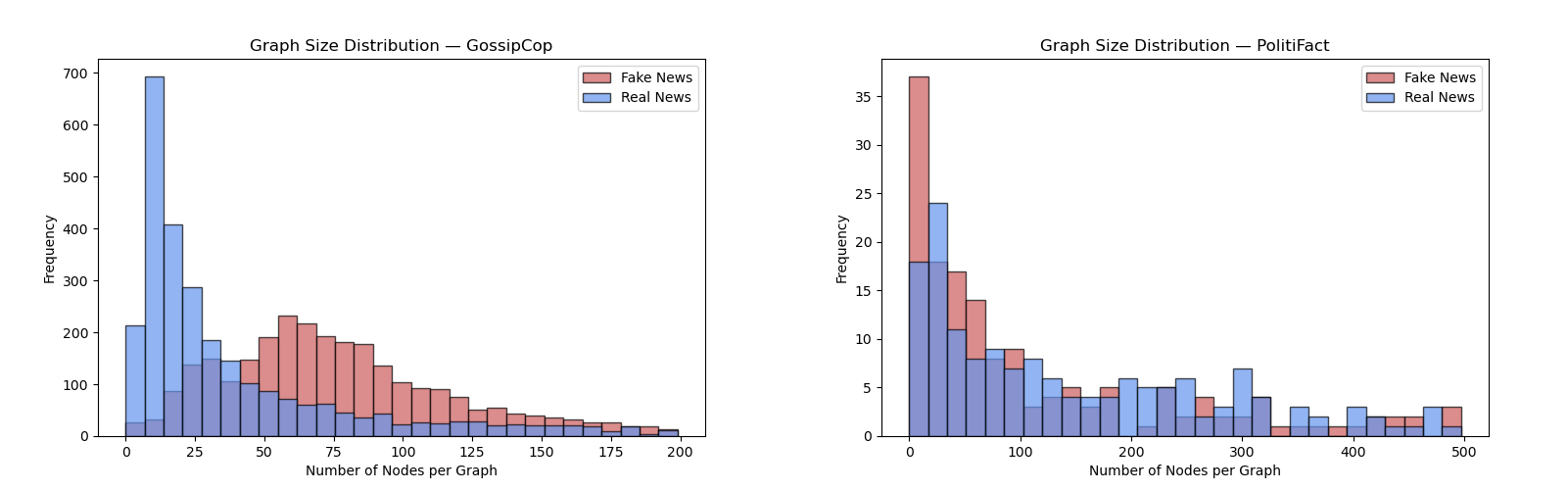}
    \caption{Graph Size Distribution for Fake and Real News in GossipCop (left) and PolitiFact (right)}
    \label{fig:graph_dist}
\end{figure}

\begin{figure}[t]
    \centering
    \includegraphics[width=1\linewidth]{images/gos_pol_shuffled.png}
    \caption{Validation accuracy over training epochs for six models (GCN, GAT, GraphSAGE, GCNFN, GNNCL, and MLP) on GossipCop (top) and PolitiFact (bottom).
Each curve corresponds to a different input setting: Base, Shuffled Features, and Shuffled Edges. 
}
    \label{fig:gos_pol_shuffled}
\end{figure}

\begin{table}[t]
\caption{Validation accuracy (mean ± 95\% CI) for GAT, GCN, GraphSAGE, GNNCL, GCNFN, and MLP on the GossipCop and PolitiFact datasets under the base setting. The best-performing model and the MLP baseline are highlighted for each dataset.}
\label{tab:base_gos-pol}
\centering
\begin{tabular}{|l|ll|}
\hline
\multicolumn{1}{|c|}{\multirow{2}{*}{Models}} & \multicolumn{2}{c|}{Datasets}                                          \\ \cline{2-3} 
\multicolumn{1}{|c|}{} & \multicolumn{1}{c|}{GossipCop} & \multicolumn{1}{c|}{PolitiFact} \\ \hline
GAT                                           & \multicolumn{1}{l|}{91.5\% ± 3.9\%}          & 82.1\% ± 4.3\%          \\ \hline
GCN                                           & \multicolumn{1}{l|}{91.2\% ± 1.7\%}          & 70.5\% ± 10.7\%         \\ \hline
GraphSAGE                                     & \multicolumn{1}{l|}{91.7\% ± 1.8\%}          & 76.3\% ± 8.6\%          \\ \hline
GNNCL                                         & \multicolumn{1}{l|}{91.2\% ± 6.3\%}          & 80.7\% ± 7.8\%          \\ \hline
GCNFN                                         & \multicolumn{1}{l|}{\textbf{96.1\% ± 1.6\%}} & \textbf{82.7\% ± 5.5\%} \\ \hline
MLP                                           & \multicolumn{1}{l|}{\textbf{90.8\% ± 3.0\%}} & \textbf{74.3\% ± 7.4\%} \\ \hline
\end{tabular}
\end{table}

\subsection{Controlled Evaluation on Synthetic Datasets}
\noindent These experiments include multiple feature regimes as described in Section 3. In addition to controlled dataset generation, we introduce a feature randomization condition which is applied only to the clean features setting where node features are shuffled independently of the graph structure. This allows us to directly test whether models are relying on structural reasoning or simply exploiting correlations between features and labels.

In the clean features setting, all models --- including the MLP --- achieve near-perfect accuracy. The node features are clean and class-separable, making the task trivially solvable without any structural information. The MLP performs as well as GNNs, confirming that expressive features alone can drive high performance. However, when we apply feature shuffling, performance drops to chance for the MLP but remains relatively stable for GNNs. This shift highlights that GNNs can fall back on structural cues, while MLPs collapse when feature-label correlation is broken.

The Noisy Features setting introduces partial corruption by flipping 20\% of node features to resemble the opposite class. This simulates misleading or noisy input. Even under this regime, GNNs outperform the MLP by leveraging structural context to override corrupted signals, indicating a strong inductive bias toward local structural consistency. 

In the structure-only setting, node features are i.i.d. noise and contain no predictive information. GNNs substantially outperform the MLP, with GAT achieving the highest accuracy. Since the MLP has no access to structure, it performs at chance. These results confirm that GNNs can learn meaningful topology-based representations in the absence of usable features.

\begin{figure}[t]
    \centering
    \includegraphics[width=0.60\textwidth]{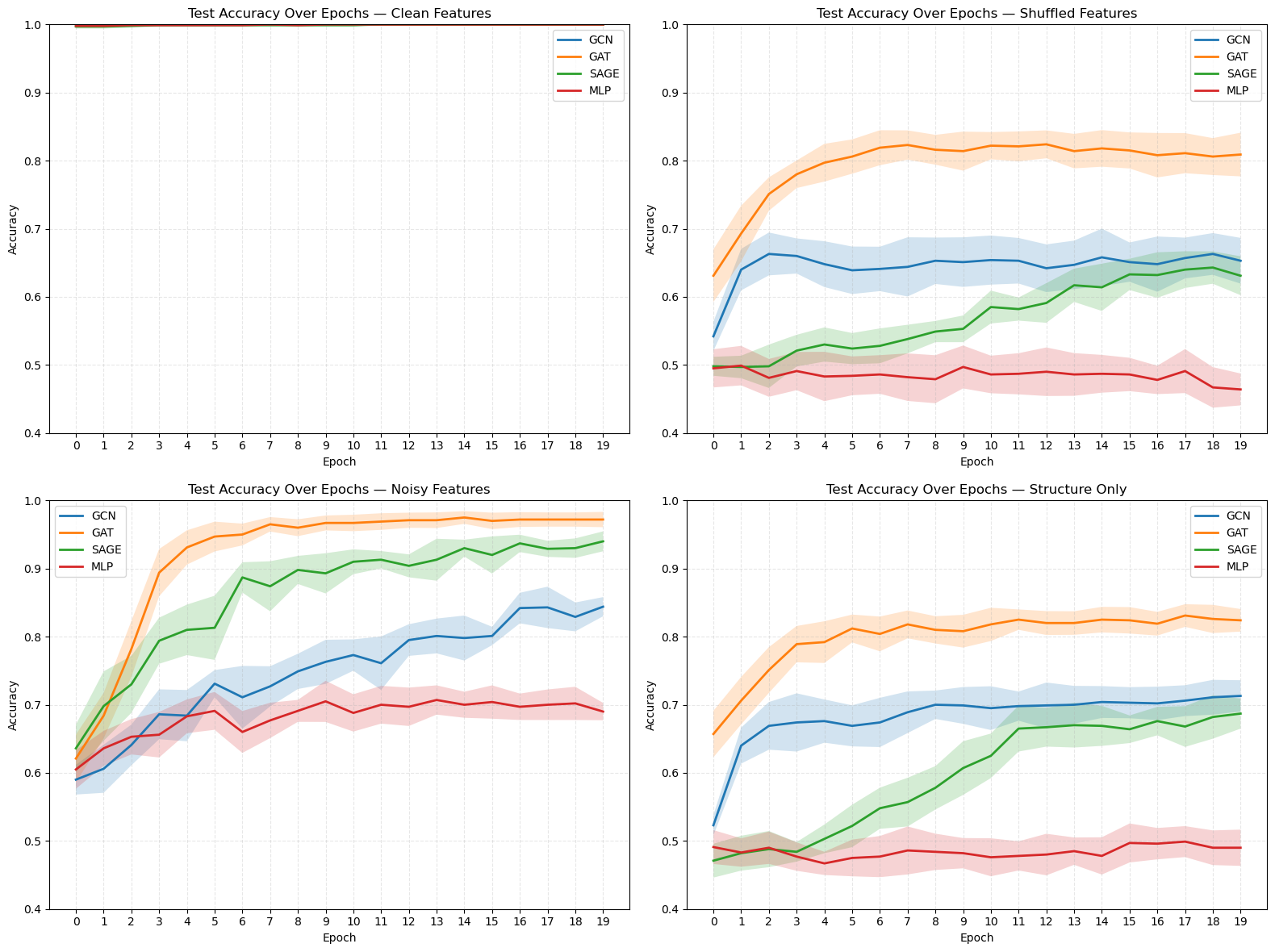}
    \caption{Test accuracy over training epochs for four models (GCN, GAT, GraphSAGE, MLP) across synthetic datasets with controlled feature and structural signals. The top row shows performance on clean separable gaussian features (left: unshuffled, right: node-randomized), where node features are highly informative. The bottom row compares performance on noisy features (left) and structure-only settings (right), where node features are misleading or absent. GNNs consistently outperform MLPs when features are weak or deceptive, while MLPs perform competitively when features are strong and class-separable.}
    \label{fig:synthetic}
\end{figure}

Taken together, these experiments demonstrate that GNNs only provide consistent benefits over structure-agnostic models like MLPs when node features are unreliable or uninformative. In settings where features are strong and consistent, such as the unshuffled clean features regime, MLPs can match or even rival GNNs. But when features are misleading or randomized, GNNs retain predictive power through structural reasoning.These trends are visualized in Figure \ref{fig:synthetic}, which shows accuracy curves across synthetic regimes.

\section{Discussion}

Our experiments reveal a significant disconnect between the structural assumptions of GNN-based fake news detection models and the realities of benchmark datasets like PolitiFact and GossipCop. Despite the widespread adoption of GNNs for this task, our findings suggest that much of their performance is attributable not to structural reasoning, but to the quality of node features.

\subsection{Informative Features Can Mask Structural Irrelevance}
Across both datasets, we observe that MLPs, which are structure-agnostic and completely ignore graph structure, achieve competitive accuracy compared to structure-aware models like GCN, GAT, GraphSAGE, GNNCL, and GCNFN. This is especially evident in GossipCop, where MLPs trail GNNs by less than 2\% accuracy, with overlapping confidence intervals. Visualizations of node features using t-SNE show linear separability between real and fake news classes, indicating that pretrained textual embeddings (e.g., BERT) alone can often solve the task.

When we perform feature shuffling, performance collapse across models --- including GAT, GCNFN and GraphSAGE --- confirming that the node features are highly discriminative. Interestingly, GCN and GNNCL retain moderate, albeit lowered accuracy under this setting in GossipCop, suggesting that they may be overfitting to residual statistical signals in the node feature distribution or relying on inductive biases that align well with spurious patterns. This challenges the notion that their robustness stems from superior structural modeling. 

These results cast doubt on the assumed structural advantage of GNNs. If strong performance is achievable with randomized structure (as shown in the shuffled edge setting) or degrades entirely when features are shuffled, then current benchmarks may overstate the role of graph topology.

\subsection{Structural Complexity is Uneven and Often Trivial}
To investigate further, we analyzed the graph topologies of both GossipCop and PolitiFact. We computed node degree distributions, degree centralization scores, and the number of 1-hop neighbors from the root node (typically the news article). These patterns are illustrated in Figure \ref{fig:normalized_root_degree}, which shows the distribution of graph sizes across real and fake news posts for both datasets.

The majority of graphs are shallow and ego-like, with over 75\% of nodes being only one hop away from the root. This limits the expressive power of neighborhood aggregation, a key mechanism for GNNs.

Notably, this structural triviality is not uniformly distributed across datasets or labels. GossipCop tends to contain highly star-shaped or fragmented structures, especially in fake news examples where propagation halts quickly. PolitiFact, by contrast, exhibits more topological diversity, including deeper trees and clustered subgraphs. This helps explain why feature shuffling impacts PolitiFact more severely: its structure is occasionally informative, and feature-label correlations are weaker. However, due to its small size, PolitiFact remains a less reliable benchmark overall, with high variance across models and perturbation settings.

These observations are visually reinforced in Figure \ref{fig:gos_pol_shuffled} and \ref{fig:synthetic}, where we plot validation curves under feature and structure perturbations for both real-world and synthetic datasets. Under edge shuffling, performance remains mostly stable and identical to base conditions, indicating that structure is not the primary driving force. Under feature shuffling, performance collapses for more than half the models tested. The few models that don't fully collapse still exhibit instability and high variance, raising questions about the kinds of signals these models are truly capturing.

\subsection{Trivial vs. Non-Trivial Structure: A Conceptual Distinction}
Many prior studies often assume that graph structure is inherently useful for GNN-based models. However, we argue that structural informativeness exists on a spectrum. In both GossipCop and PolitiFact, most graphs lean toward the trivial end.

Trivial graphs are typically centered around a single root node, with nearly all other nodes connected directly to it. These structures resemble star or ego-graphs and do not meaningfully benefit from message passing. In these cases, models like MLPs perform nearly as well as GNNs.

\begin{figure}[t]
  \centering
  \includegraphics[width=0.48\linewidth]{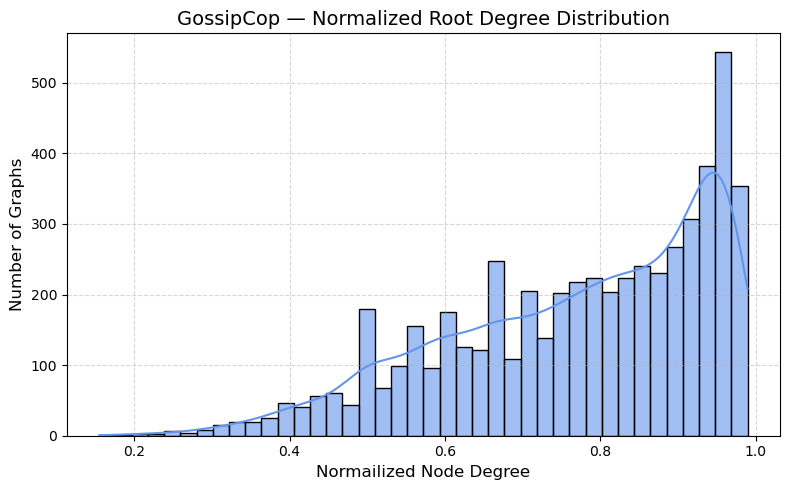}
  \includegraphics[width=0.48\linewidth]{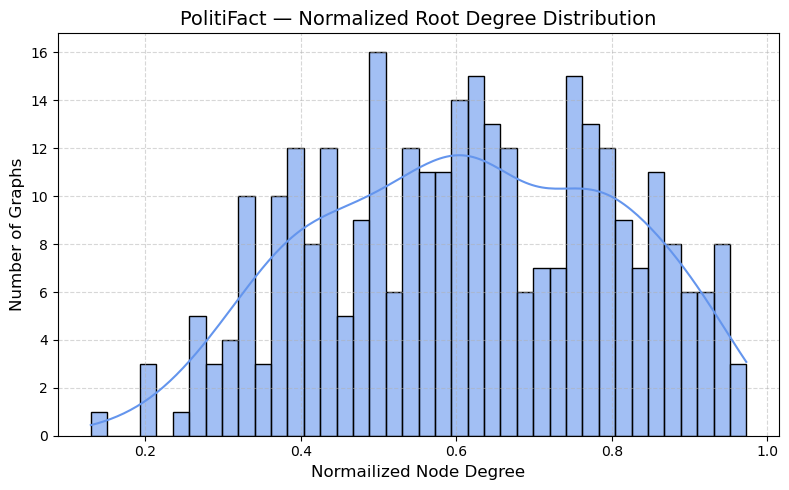}
  \caption{Normalized root degree distributions for GossipCop (left) and PolitiFact (right).  
  Values closer to 1 indicate ego-like star graphs where most nodes connect directly to the root post.  
  GossipCop exhibits more extreme centralization across its examples.}
  \label{fig:normalized_root_degree}
\end{figure}

To quantify this centralization, we compute the normalized degree of the root node, defined as the number of edges connected to the root divided by the maximum possible degree in a connected graph of the same size. Formally:
\[
\text{Normalized Root Degree} = \frac{\deg(\text{root})}{N - 1}
\]
where $deg(root)$ is the degree of the root node (typically the original news article), and $N$ is the number of nodes in the graph. This metric ranges from 0 to 1 and captures the extent to which a graph exhibits a star-like or ego structure. High values indicate that most nodes are directly connected to the root, while lower values suggest a more distributed topology.  The distribution of normalized root degrees across both datasets is shown in Figure~\ref{fig:normalized_root_degree}, illustrating that GossipCop graphs are more centralized than those in PolitiFact. 

In GossipCop, most graphs have between 5 and 75 direct neighbors, with some extending up to 190+. In PolitiFact, the distribution is similar but more extreme, with some graphs exceeding 400 1-hop connections. These patterns reflect very high degree centralization and indicate that message passing is largely constrained to a single hop. The summary statistics for normalized root degree are reported in Table \ref{tab:normalizedrootdeg}.

\begin{table}[t]
\caption{Summary Statistics for Normalized Root Degree across GossipCop and PolitiFact}
\label{tab:normalizedrootdeg}
\centering
\begin{tabular}{|l|l|l|l|}
\hline
Dataset    & Mean  & Median & Std. Dev. \\ \hline
GossipCop  & 0.771              & 0.805                & 0.169     \\ \hline
PolitiFact & 0.614              & 0.617                & 0.189     \\ \hline
\end{tabular}
\end{table}

These distributions confirm that GossipCop graphs are heavily centralized, with the majority of edges concentrated around a single hub. PolitiFact graphs exhibit slightly more structural variation, but still lean toward high centralization. 

We emphasize that normalized root degree captures a continuous notion of centralization rather than a binary label. Instead of asking whether a graph is ego-like or not, it is more informative to ask: \textit{“To what degree is a graph structure informative?”} Low-centrality graphs exhibit more branching, clustering, or intermediate hubs — properties that enable multi-hop message passing. These topologies can make structure genuinely useful, especially when node features are noisy or less informative. Our synthetic results confirm that GNNs only provide a performance advantage in such cases.

\section{Next Steps}

Our findings raise important concerns about the evaluation and interpretation of GNN-based fake news detection. While our focus was on evaluating benchmarks and structural analysis, we believe these insights point towards three concrete next steps for improving research in this area:  developing better datasets, evaluating structural properties before use, and reconsidering modeling methodology.

\subsection{Seek Out Datasets with Richer Structure}
Current benchmarks like GossipCop and PolitiFact reward models that rely heavily on node features rather than structural reasoning. Future datasets should include greater topological diversity, such as multi-hop propagation chains, intermediary hubs, or community-based spreading, that meaningfully test a model’s ability to leverage structure. Incorporating temporal dynamics (e.g., the evolution of news over time) would further simulate realistic propagation and challenge static graph assumptions.

Promising efforts like like MuMiN \cite{nielsen2022mumin} point in this direction. MuMiN includes multi-modal features (e.g., text, images, hashtags), real user-claim interactions, and multilingual content collected over more than a decade. Evaluating GNNs on such data could offer a better understanding of whether structure contributes meaningfully to generalization, particularly in more dynamic and noisy real-world scenarios.

\subsection{Evaluate Datasets Before Use}
Researchers should not treat benchmark datasets as structurally informative without validation. Structural informativeness varies across datasets and should be empirically measured. Metrics such as normalized root degree, depth distributions, and clustering coefficients can quantify how much structure is present and how useful it may be. We recommend that structural analysis become a standard practice before using GNNs, similar to how class imbalance or feature leakage are now routinely checked in classical ML pipelines.

\subsection{Rethink Methodology}
If node features alone suffice to solve the task, GNNs may add unnecessary complexity. Models should only incorporate message passing when there is evidence that structure contributes predictive value. Future research should isolate and quantify the marginal contribution of structure, possibly through ablation studies, synthetic controls, or feature-structure disentanglement frameworks. Additionally, incorporating structural bias only when warranted  could help prevent overfitting to trivial graph topology.

\section{Conclusion}

In this work, we revisited the use of graph neural networks for fake news detection by critically examining the structural and feature-based assumptions embedded in two widely used benchmarks: GossipCop and PolitiFact. Through systematic experiments on both real-world and synthetic datasets, we found that the strong performance of GNNs on these benchmarks is largely attributable to the informativeness of node features, particularly those derived from pretrained language models, rather than meaningful structural signals.

Across both datasets, a simple MLP, completely agnostic to graph structure, achieved performance within 1-2\% of GCN, GAT, GraphSAGE, and GCNFN, often with overlapping confidence intervals. On GossipCop, for instance, the MLP reached 93.87\% accuracy compared to 95.26\% for GAT and 95.54\% for GraphSAGE. Even GCNFN, which combines GAT layers with fused textual content, only marginally outperformed the MLP. These results suggest that structural signals contribute little when node features are already highly discriminative.

Our controlled synthetic experiments further support this conclusion. When node features are Gaussian-distributed and class-separable, all models, including MLPs, achieve near-perfect accuracy. But when features are noisy or randomized, GNNs outperform MLPs by up to 20\%, leveraging structure to maintain predictive performance. This contrast highlights that GNNs are only truly effective when structure plays a necessary disambiguating role.

\section*{Acknowledgments}
This section will be added in the camera-ready version.

\bibliography{software}

\end{document}